\documentclass[10pt,twocolumn,letterpaper]{article}

\usepackage[accsupp]{axessibility}  
\usepackage{iccv}
\usepackage{times}
\usepackage{epsfig}
\usepackage{graphicx}
\usepackage{amsmath}
\usepackage{multirow}
\usepackage{multirow}
\usepackage[font={small}]{caption}
\usepackage{ifthen}
\usepackage{bm}
\usepackage{subcaption}
\usepackage{balance}
\usepackage{multirow}
\usepackage{lipsum}
\usepackage{booktabs}
\usepackage{verbatim}
\usepackage{graphicx}
\usepackage{bbding}
\usepackage{pifont}
\usepackage{wasysym}
\usepackage{diagbox}
\usepackage{ifthen}
\usepackage{bbding} 
\usepackage{bm}
\usepackage{threeparttable}
\usepackage{lipsum}
\usepackage{animate}
\usepackage{rotating} 
\usepackage{makecell}
\usepackage{amsmath}
\usepackage{amssymb}
\usepackage{bbding}
\usepackage{pifont}
\usepackage{lipsum}
\usepackage{lipsum}
\usepackage{algorithmicx}
\usepackage{algorithm}
\usepackage{algpseudocode}
\usepackage{amsmath}

\usepackage{textcase}

\usepackage{textcomp,booktabs}

\usepackage[pagebackref=true,breaklinks=true,letterpaper=true,colorlinks,bookmarks=false]{hyperref}

\iccvfinalcopy 

\def\iccvPaperID{1212} 

\begin{document}

\setlength{\abovedisplayskip}{3pt}
\setlength{\belowdisplayskip}{3pt}

\title{Generalized Lightness Adaptation with Channel
Selective Normalization}

\author{Mingde Yao$^1$\footnotemark[1], Jie Huang$^1$\footnotemark[1], Xin Jin$^2$, Ruikang Xu$^1$, Shenglong Zhou$^1$, Man Zhou$^3$, Zhiwei Xiong$^1$$^\dag$\\
$^1$University of Science and Technology of China~~~~~~~ \\$^2$Eastern Institute of Technology\\$^3$Nanyang Technological University \\
}
\maketitle

\renewcommand{\thefootnote}{\fnsymbol{footnote}} 
\footnotetext[1]{Co-first author.}
\footnotetext{$^\dag$Corresponding author: zwxiong@ustc.edu.cn.}


\begin{abstract}

Lightness adaptation is vital to the success of image processing to avoid unexpected visual deterioration, which covers multiple aspects, e.g., low-light image enhancement, image retouching, and inverse tone mapping. Existing methods typically work well on their trained lightness conditions but perform poorly in unknown ones due to their limited generalization ability. To address this limitation, we propose a novel generalized lightness adaptation algorithm that extends conventional normalization techniques through a channel filtering design, dubbed Channel Selective Normalization (CSNorm). The proposed CSNorm purposely normalizes the statistics of lightness-relevant channels and keeps other channels unchanged, so as to improve feature generalization and discrimination. To optimize CSNorm, we propose an alternating training strategy that effectively identifies lightness-relevant channels. The model equipped with our CSNorm only needs to be trained on one lightness condition and can be well generalized to unknown lightness conditions. Experimental results on multiple benchmark datasets demonstrate the effectiveness of CSNorm in enhancing the generalization ability for the existing lightness adaptation methods. Code is available at \href{https://github.com/mdyao/CSNorm/}{https://github.com/mdyao/CSNorm}. 

\end{abstract}

\section{Introduction}

Lightness adaptation is a vital step in image processing, encompassing tasks such as low-light image enhancement~\cite{li2018structure,liu2021retinex}, image retouching~\cite{liang2021ppr10k}, 
and inverse tone mapping~\cite{chen2021new}. These tasks have benefited significantly from the development of advanced neural network architectures. Although numerous powerful lightness adaptation methods have been proposed, the generalization problem~\cite{zedusr,seo2020learning} for lightness adaptation still exists and is rarely explored.

\begin{figure}[!t]
\centering
\includegraphics[width=0.49\textwidth]{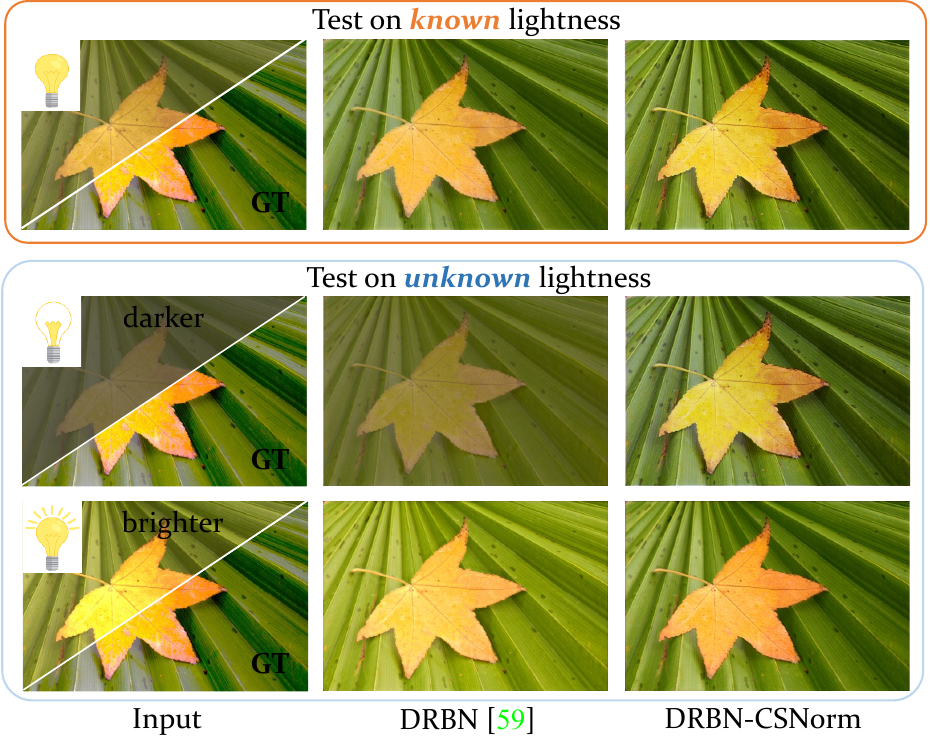}
\vspace{-0.2in}
    \caption{Visual comparisons on \emph{known} and \emph{unknown} lightness conditions. The model equipped with our CSNorm can generalize well to other \emph{unknown} lightness while keeping the performance on the \emph{known} lightness. }
\label{fig:intro1_light}
\vspace{-0.2in}
\end{figure}

In real-world applications, applying lightness adaptation models to unknown lightness conditions is quite challenging due to the brightness discrepancies between training and testing data~\cite {ni2020towards,wang2022ucl}. Existing lightness adaptation approaches~\cite{liu2021retinex,zhang2019kindling,yang2020fidelity,lore2017llnet,afifi2021learning,song2021starenhancer,chen2021new,kim2019deep} primarily focus on addressing the challenge of accurate image reconstruction.
However, they often underperform on wide-range scenes with other lightness conditions due to their over-fitting to the training lightness component, leading to unsatisfactory visual effects (Fig.~\ref{fig:intro1_light}) and  inadequate generalization in complex real-world scenarios.

An alternative way is constructing a larger mixed-lightness dataset including more lightness conditions, but it is impractical for many complicated cases and too time-consuming for cumbersome acquisition from diverse domains~\cite{shankar2018generalizing,zhou2020learning}. Besides, existing models suffer from the drawback of inadequately encapsulating generalization and discrimination abilities, where the former is responsible for the performance on unknown lightness conditions and the latter mainly corresponds to the reconstruction characteristics on the known lightness condition.

\begin{figure}[!t]
\centering
    \includegraphics[width=0.98\linewidth]{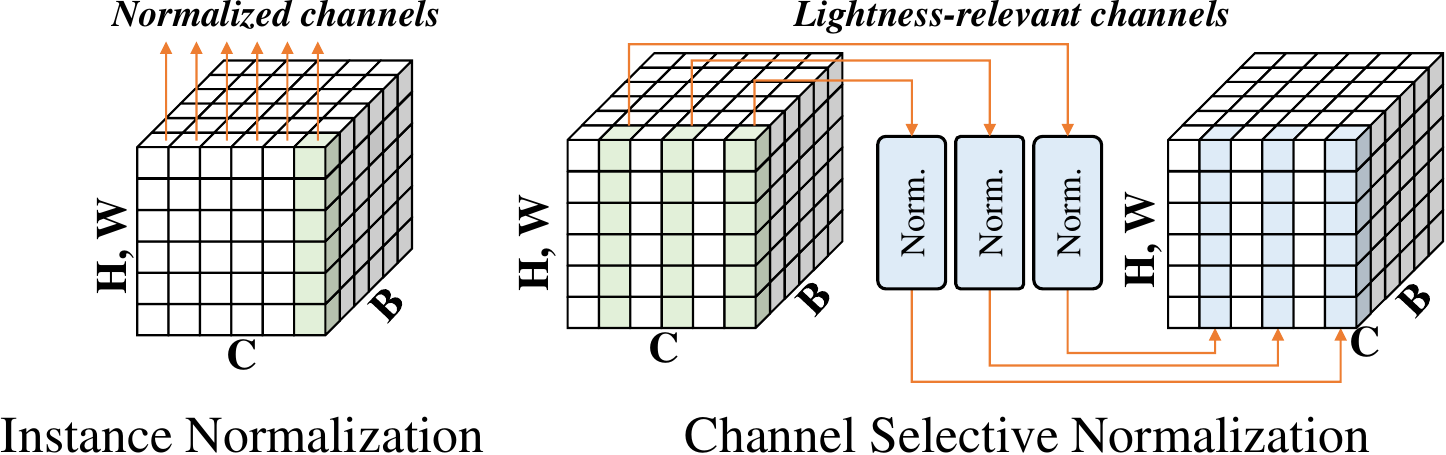}
    \vspace{-0.07cm}
    \caption{Comparisons of different normalization techniques. Our proposed CSNorm adaptively normalizes the lightness-relevant channels for generalization and keeps other channels unchanged for accurate reconstruction. }
\label{fig:intro1_norm}
\vspace{-0.05in}
\end{figure}

In this paper, we focus on designing a mechanism that empowers existing lightness adaptation methods with the generalization ability to wide-range lightness scenes. 
The key challenge lies in obtaining the above generalization ability while keeping the discrimination ability. 
To achieve this goal, we introduce the normalization technique, which has the good property of extracting invariant representations from the given features~\cite{jin2020style}, especially for lightness components~\cite{pan2018two,huang2022exposure}. 
However, normalization is a \emph{double-edged sword} due to its inevitable loss of information, which might degrade the reconstruction accuracy~\cite{pan2018two,zhou2022normalization}. Therefore, we explore normalizing particular channels that are highly sensitive to lightness changes while keeping other channels unchanged (Fig.~\ref{fig:intro1_norm}). 
Such a design enhances the generalization ability and keeps the discrimination of features.

To this end, we propose a concept of \emph{Channel Selection Normalization} (CSNorm) to purposely select and normalize the lightness-relevant channels. CSNorm consists of two major parts: an instance-level lightness normalization module for eliminating lightness-relevant information and a differentiable gating module for adaptively selecting lightness-relevant channels. The gating module outputs a series of binary indicators to combine the normalized and original channels, which feasibly enhances the model’s generalization capabilities and mitigates the information loss caused by normalization.
The proposed CSNorm is simple, lightweight, and plug-and-play.

To identify lightness-relevant channels in CSNorm, we meticulously design an alternating training strategy. The network is alternately optimized with different inputs of two steps. Specifically, in the first step, the network inputs the images to learn an essential ability for lightness adaptation. In the second step, we slightly perturb the lightness of the above input image and solely optimize CSNorm with other parameters frozen. Since the only variable in the input images is the lightness condition, CSNorm can adaptively identify lightness-relevant channels and normalize them accordingly, thereby exhibiting superior performance in terms of generalization and discrimination.

In summary, we make the following contributions.

{$\bullet$} To our best knowledge, this is the first work that improves the generalization ability of lightness adaptation methods in wide-range lightness scenarios.

{$\bullet$} We propose CSNorm, which selectively normalizes the lightness-relevant channels according to their sensitivity to lightness changes. The model equipped with our CSNorm can generalize well to unknown lightness conditions while keeping the reconstruction ability on known lightness conditions.

{$\bullet$} An alternate training strategy is meticulously designed to effectively optimize CSNorm for identifying lightness-relevant channels.

{$\bullet$} We conduct extensive experiments to validate the advantage and versatility of CSNorm over existing lightness adaptation methods for improving their generalization in wide-range lightness scenarios.

\section{Related Work}
\label{Sec:Related Work}

\subsection{Lightness Adaptation}

As a key step in image restoration~\cite{zhao2023ddfm,pan2022towards,fu2021model,yang2022memory,luo2023effectiveness}, lightness adaptation tasks~\cite{lightadapt,jin2022unsupervised,zhang2022structure,huang2022exposure1,afifi2021learning}, such as low-light image enhancement~\cite{li2018structure,liu2021retinex,zhang2022structure,zhang2019kindling,yang2020fidelity,lore2017llnet,ma2022toward}, image retouching~\cite{liang2021ppr10k,zeng2023region}, 
and inverse tone mapping~\cite{mantiuk2008display,chen2021new,yao2023bidirectional,kim2019deep,yao2023bidirectional}, aim to adjust lightness components (e.g., illumination, color, and dynamic range) from a degraded version to a normal version. 
Low-light image enhancement aims to improve the visibility and quality of images captured under low-light conditions. In recent years, various deep learning-based approaches~\cite{liu2021retinex,zhang2019kindling,yang2020fidelity,lore2017llnet,ren2020lr3m} have shown promising results in this field.
For image retouching, CSRNet~\cite{he2020conditional} formulates pixel-independent operations by multi-layer perceptrons (MLPs), which learns implicit step-wise retouching operations.
Inverse tone mapping aims to translate images from high dynamic range to low dynamic range. HDRTVNet~\cite{chen2021new} proposes a multi-stage method to adjust the global intensity and local contrast step-by-step. SR-ITM~\cite{kim2019deep} proposes a dynamic filter to jointly learn the super-resolution and inverse tone mapping with a single network. 
Although prior works have made significant progress on lightness adaptation, they inherently tend to overfit the training data, resulting in poor generalization performance. 
Our proposed CSNorm enables the model to generalize to unknown lightness conditions, which only needs to be trained with limited lightness conditions and avoids time-consuming data collection. 
Besides, its lightweight and plug-and-play nature allows for easy integration into various networks.

\subsection{Generalization}

A well-generalized model exhibits the ability to infer meaningful patterns, relationships, and features from its training data and apply them effectively to new, unseen instances. Various methods have been developed to address this problem, including domain generalization~\cite{seo2020learning,chattopadhyay2020learning}, self-supervised learning~\cite{yaoselfdn,lirui1,feng2022docgeonet}, unsupervised learning~\cite{yuanbiao1}, contrastive learning~\cite{yuanbiao2}, and zero-shot learning~\cite{zedusr}. In this paper, we focus on the generalization problem of lightness adaptation tasks. 
Domain 
generalization~\cite{seo2020learning,chattopadhyay2020learning,huang2020self} aims to learn the domain-invariant representation from multiple source domains which could generalize well on unseen domains. Existing methods tried to address it mainly from the dataset synthesis aspect~\cite{shankar2018generalizing,zhou2020learning} or optimization algorithm aspect~\cite{li2018learning,dou2019domain,du2020learning}. 
Beyond domain generalization, single domain generalization~\cite{qiao2020learning,fan2021adversarially} has gained interest recently. This task aims to learn the model from one source domain to get a well generalization ability on other unseen domains. Following the development trajectory of domain generalization, methods based on adversarial domain augmentation have been proposed. 
However, our proposed method is distinct from existing approaches in that it is simple, lightweight, and specifically designed to meet the lightness adaptation requirements.

\begin{figure}[t]
  \centering
  \includegraphics[width=1\linewidth]{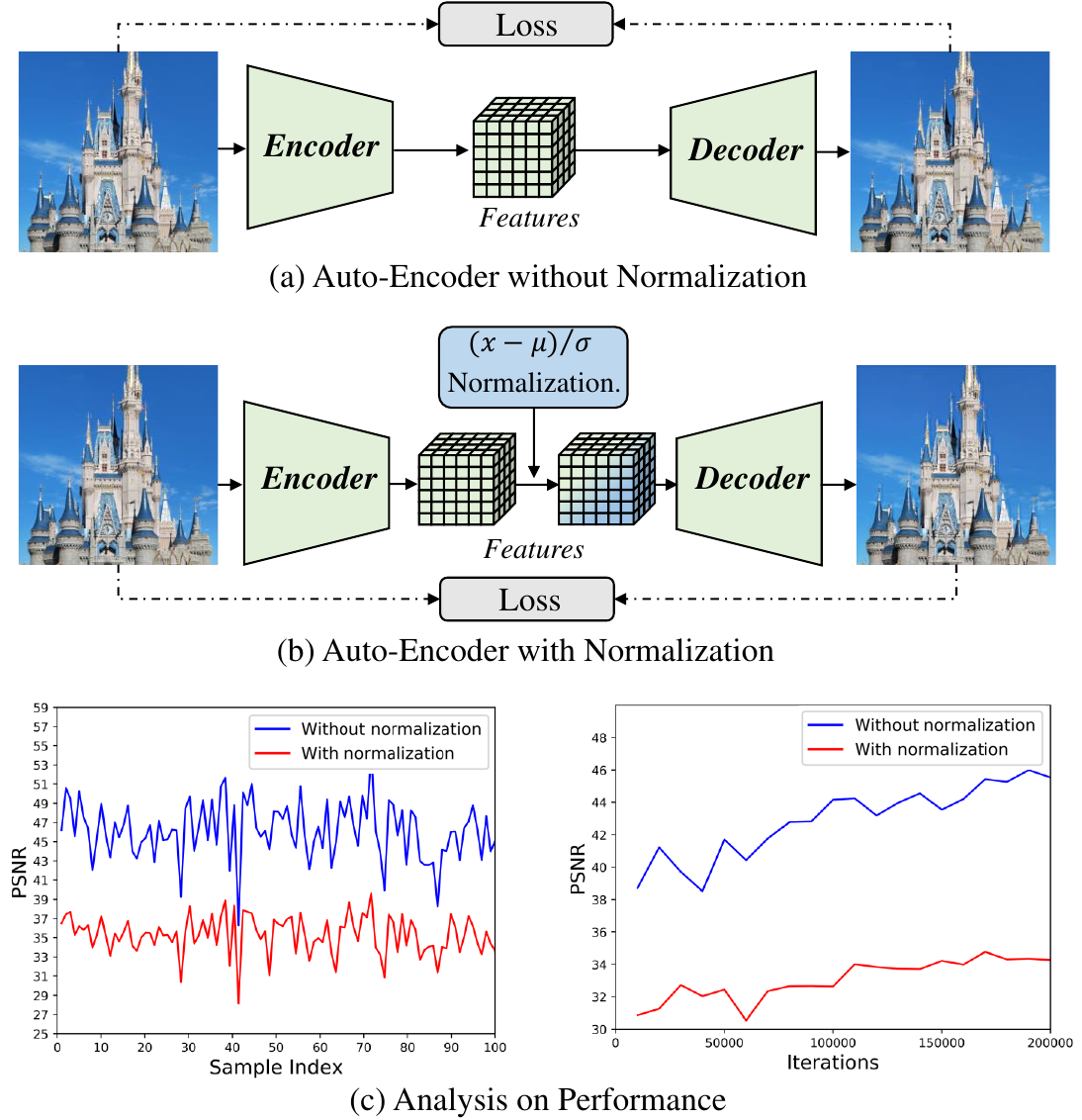}
  \vspace{-0.5cm}
  \caption{The effect of applying normalization for image reconstruction. As can be seen, the introduction of instance normalization is harmful to image reconstruction.}\label{fig:sae}
  \vspace{-0.1in}
\end{figure}

\subsection{Normalization}

Normalization plays an essential role in image processing, especially for lightness-relevant tasks. Formally, normalization subtracts the mean value and is derived by standard to scale the image in classical image processing.
In deep learning-based methods, normalization serves as a basic layer~\cite{ioffe2015batch,wu2018group,ulyanov2016instance,ba2016layer} with various varieties. 
Recently works~\cite{fan2021adversarially,skorokhodov2020class,segu2020batch,nam2018batch} point that normalization has a good property of generalization for neural networks. BN-Test~\cite{nado2020evaluating} calibrats the model under covariate shift at the test stage. ASR-Norm~\cite{fan2021adversarially} adapts to each individual input sample to avoid dependency on the testing samples. Though normalization has been investigated in high-level vision, it is rarely explored in the low-level lightness adaptation field.

\begin{figure*}[t]
    \centering
    \includegraphics[width=0.98\linewidth]{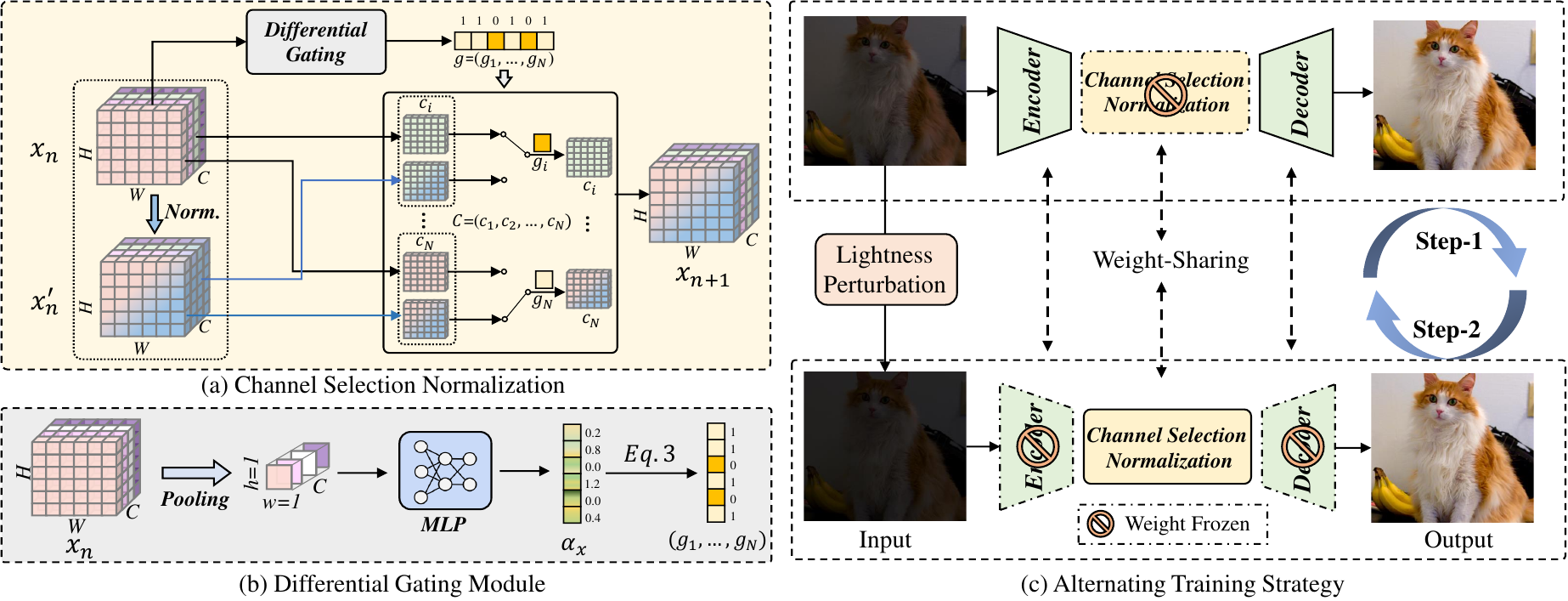}
    \vspace{-0.2cm}

    \caption{Overview of our proposed method. (a) Channel selective normalization (CSNorm), which consists of an instance-level normalization module and a differential gating module. 
    (b) Differential gating module. It outputs a series of on-off switch gates for binarized channel selection in CSNorm. (c) Alternating training strategy. In the first step, we optimize the parameters outside CSNorm to keep an essential ability for lightness adaptation. In the second step, we only update the parameters inside CSNorm (see (a)\&(b)) with lightness-perturbed images. The two steps drive CSNorm to select channels sensitive to lightness changes, which are normalized in $x_{n+1}$. }
    
    \label{fig:main_pipeline}
\vspace{-0.2cm}
\end{figure*}

\vspace{-0.1cm}
\section{Motivation}
\label{headings}
\vspace{-0.1cm}

Since lightness differs and varies substantially in real-world captures, the processing of the lightness adaptation method is significantly variable. Consequently, it is challenging to directly deploy existing networks for real-world scenarios, particularly in lightness conditions absent from the training set. An alternative way is to increase the dataset's capacity by creating an enlarged mixed dataset, including extra lightness conditions for training. However, the exorbitant cost of the data collection makes it a challenging proposition to pursue. Furthermore, the mixed dataset has a greater propensity to cause ambiguity during training, which might bias the network toward particular lightness and result in an imbalanced training issue~\cite{xiao2021improving}.

Normalization has good properties of eliminating lightness-relevant components~\cite{huang2022exposure} and reducing the discrepancy between images~\cite{muandet2013domain}. It can effectively lessen the impact of lightness and competently extract lightness-independent information, which enables the network to learn robust representations and improve the generalization ability. Based on this point, we aim to present a general normalization algorithm to address the generalization problem for lightness adaptation. 

Despite these benefits, normalization is a \emph{double-edged sword} for networks due to the inevitable loss of information (\emph{e.g.}, statistical characteristics including mean and variance)~\cite{pan2018two,zhou2022normalization}, resulting in inferior reconstruction performance. 
To comprehend the influence of normalization intuitively, we conduct a self-reconstruction task to  illustrate the information loss  induced by normalization.
As shown in Fig.~\ref{fig:sae}, we train two auto-encoder networks separately with and without inserting the IN~\cite{ulyanov2016instance} operation and we calculate the relative reconstruction accuracy (\emph{i.e.,} PSNR) on different images. It is obvious that normalization ruins the network's reconstruction ability, and in fact, the harm of information loss caused by normalization outweighs its potential benefits in terms of generalization. This motivates us to design CSNorm to selectively normalize the channels, concurrently considering the generalization ability and reconstruction accuracy for lightness adaptation.

\vspace{-0.1cm}
\section{Method}
\vspace{-0.1cm}
\subsection{Overview} 
Based on the above analysis, we propose a simple yet effective method as shown in Fig.~\ref{fig:main_pipeline}. Particularly, we design CSNorm (Fig.~\ref{fig:main_pipeline}\textcolor{red}{a}) to improve the network's capacity for generalization, which can be used as a plug-and-play module for existing lightness adaptation networks. In CSNorm, 
a differentiable gating module (Fig.~\ref{fig:main_pipeline}\textcolor{red}{b}) is introduced to efficiently select the original and normalized features along the channel dimension and then combine them to be passed to the next layer. Such a gating module operates as a de facto channel-selection function. 

Further, we propose an alternating training strategy to force the gating module to select lightness-relevant channels, which is driven by performance stability under lightness perturbations (Fig.~\ref{fig:main_pipeline}\textcolor{red}{c}). During the training stage, we only access one dataset with limited lightness conditions and train the model equipped with our CSNorm. Once trained, the model can work directly on other unknown lightness conditions.

\subsection{Channel Selective Normalization}

As depicted in Fig.~\ref{fig:main_pipeline}\textcolor{red}{a}, CSNorm consists of two  parts: an instance-level lightness normalization module for eliminating light-relevant information and a differentiable gating module for adaptively selecting light-relevant channels.

\subsubsection{Instance-level Lightness Normalization}
To facilitate the subsequent selection of channels, we normalize the channels and adopt IN as the implementation to operate precisely on individual instances and channels.
Given a feature $x$ with the shape of $H\times W \times C$, IN normalizes $x$ by subtracting the mean value $\mu(x)$ followed by dividing the standard deviation $\sigma(x)$, which is expressed as
\begin{equation}\label{eq:IN}
    x'={\rm IN}(x)=\gamma \frac{{x}-\mu(x)}{\sigma(x)}+\beta,
\end{equation}
where $\mu(x)$ and $\sigma(x)$ are calculated independently across spatial dimensions for each channel and instance, and $\gamma, \beta\in \mathbb{R}^C$ are scalable parameters learned from data. Since IN can reduce the lightness discrepancy among instances~\cite{jin2020style}, the normalized feature $x'$ has a robust representation irrelevant to the lightness conditions, enabling the network to adapt to various lightness scenarios and improving its generalization capability. 

\subsubsection{Differentiable Gating Module}
To achieve adaptive channel selection with minimum network modification costs, we introduce a differentiable gating module for channel selection, which feasibly enhances the model's generalization capabilities and mitigates the information loss caused by normalization. As depicted in Fig.~\ref{fig:main_pipeline}\textcolor{red}{a}, the differentiable gating module outputs a series of binary indicators to combine the normalized and original channels,  which can be expressed as
\begin{equation}\label{eq:gate_interp}
\setlength{\abovedisplayskip}{3pt}
x_{n+1} = (\mathbf{1}-g)\odot x_{n}+g\odot x'_n.
\setlength{\belowdisplayskip}{3pt}
\end{equation}
where $g$ represents the binary indicators across the channel dimension, and $\odot$ is the channel-wise multiplication. The gating operation activates or deactivates the channels by the binary indicators to normalize channels selectively. Consequently, the generated feature $x_{n+1}$ eliminates the effects of lightness to obtain an invariant representation for generalization, and retains the essential information with unchanged channels for accurate reconstruction.

Specifically, the gating operation is expected to be differentiable and capable of biasing the output to zero or one for channel selection. Inspired by the pruning methods sampling the filters~\cite{you2019gate}, we construct the gating module as
\begin{equation}\label{eq:gating function}
\setlength{\abovedisplayskip}{1pt}
g=G(\alpha_x)=\frac{\alpha^2_x}{\alpha^2_x+\epsilon},
\setlength{\belowdisplayskip}{1pt}
\end{equation}
where $\alpha_x\in \mathbb{R}^C$ is an intermediate vector generated from the feature $x$, and $\epsilon$ is a small positive number. Specifically, to obtain $\alpha_x$, we first employ adaptive pooling to shrink the spatial size of $x$ to a single pixel, followed by several fully-connected layers with ReLU activations (Fig.~\ref{fig:main_pipeline}\textcolor{red}{b}).

When $\alpha_x=0$, it is obvious that $G(0)=0$; when $\alpha_x \neq 0$, we can infer that $G(\alpha_x)\approx 1$ since $\epsilon$ is small enough. This function transforms $\alpha_x$ to a value close to one or zero, resulting in an on-off switch gate without requiring additional manual threshold design. Moreover, leveraging its differentiable character, we design an 
alternating optimization strategy (Sec.~\ref{sec:joint training}) to adaptively select the lightness-relevant channels. 

It is worth noting that, the gating module can easily fall into a trivial solution that keeps all the channels unchanged to preserve the reconstruction accuracy, since the output of the function can easily be one (when $\alpha_x\neq 0$ ). Consequently, we make $g$ to directly multiply with normalized channels $x'$ in Eq.~\ref{eq:gate_interp}, pushing the network to prefer normalized channels to keep an elegant balance between the model's generalization ability and the discrimination of features for reconstruction.

\subsection{Alternating Training Strategy}\label{sec:joint training}

\subsubsection{Training Strategy}

We propose an alternating training strategy, as illustrated in Fig.~\ref{fig:main_pipeline}\textcolor{red}{c}, to locate lightness-relevant channels in CSNorm. The rationale behind our strategy is that, by slightly perturbing the lightness condition of the input images, CSNorm is forced to locate and filter out lightness-relevant channels to achieve optimal performance on both original and perturbed images. Specifically, the strategy alternately optimizes the network on the original dataset to learn an essential ability for lightness adaptation, and optimizes CSNorm with slightly perturbed input images to identify lightness-relevant channels. This ensures that CSNorm can efficiently normalize lightness-relevant channels, exhibiting superior performance in both generalization and discrimination.

\begin{figure}[t]
    \centering
    \includegraphics[width=0.98\linewidth]{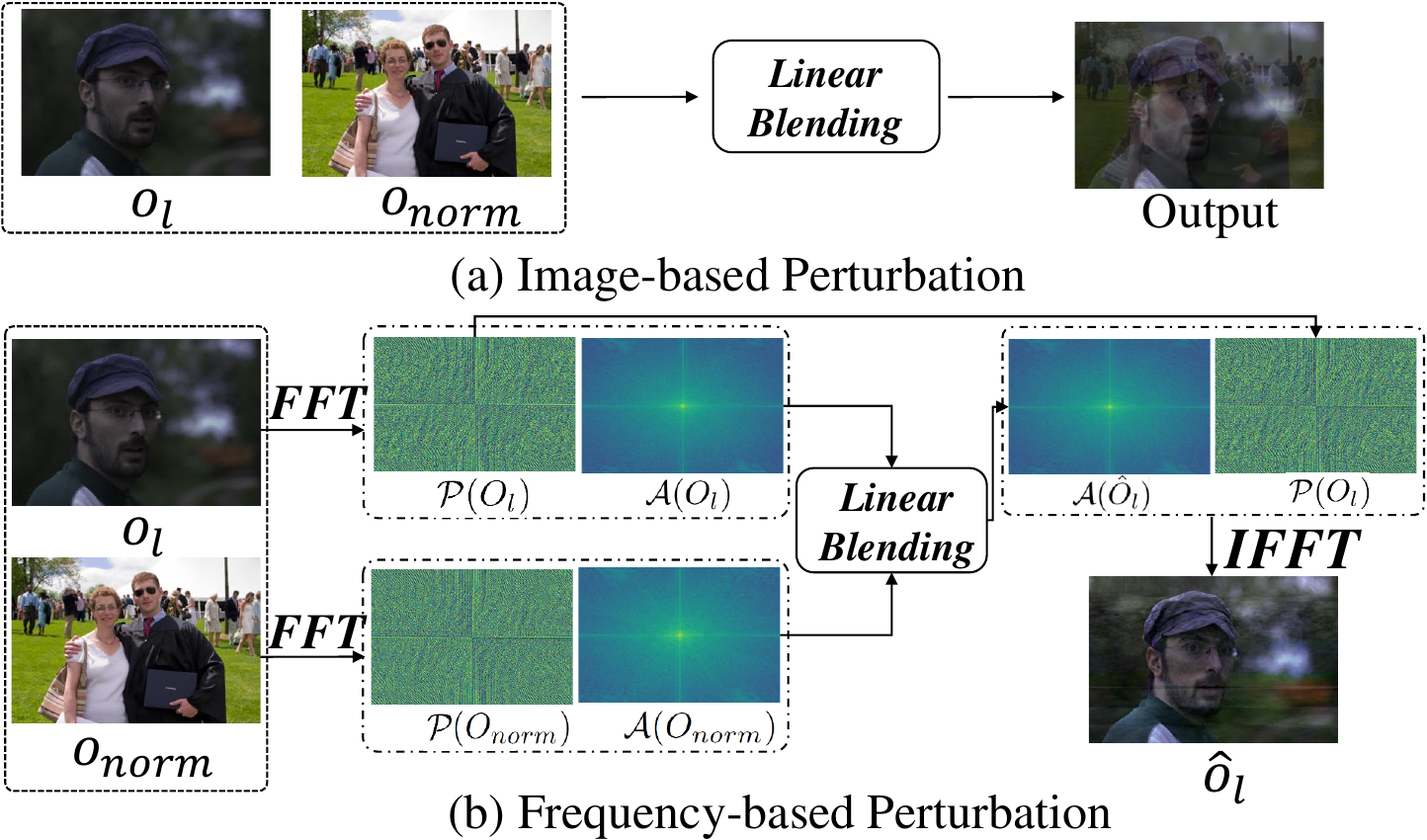}
    \vspace{-0.2cm}\caption{Comparisons of lightness perturbations. Our frequency-based perturbation captures the essential component of lightness, while avoiding interfering with other image components such as structural information.}
    \label{fig:frequency_aug}
    \vspace{-0.3cm}
\end{figure}

To optimize the network, we separate its parameters into two groups based on whether they belong to CSNorm, and use different loss functions to update them, as shown in Fig.~\ref{fig:main_pipeline}\textcolor{red}{c}. In the first step, we  input the original image and update parameters outside CSNorm by minimizing the loss function \begin{equation}
\mathcal{L}_{1}=|\hat{o}_1-o_{gt}|_2,
\end{equation}
where $\hat{o}_1$ is the output image of the network and $o_{gt}$ is the ground-truth image. By doing so, the network's essential lightness-adaptation capability is maintained and all channels are preserved in their natural state. 

In the second step, we perturb the lightness of the input image (Sec.~\ref{sec:lightness}) and fix the parameters outside CSNorm. In other words, we only update the parameters in CSNorm, by minimizing the loss function 
\begin{equation}\label{eq:trainigloss2}
\mathcal{L}_{2}=|\hat{o}_2-o_{gt}|_2+|\mathcal{A}(\hat{o}_2)-\mathcal{A}(o_{gt})|_2,
\end{equation}
where $\hat{o}_2$ is the output, and $\mathcal{A}$ reprensents the amplitude information in frequency domain. This enables CSNorm to adaptively select the lightness-relevant channels to keep the performance on perturbed images. In particular, since lightness is related to magnitude in the frequency domain~\cite{huangdeep2022}, we add the amplitude loss $|\mathcal{A}(\hat{o}_2)-\mathcal{A}(o_{gt})|_2$  in Eq.~\ref{eq:trainigloss2} to allow the network to focus more on lightness information and effectively select lightness-relevant channels.

The two steps are alternately optimized by the above two objectives, and the overall optimization function is given by
\begin{equation}
\mathcal{L}=\mathcal{L}_{ori}+\delta \mathcal{L}_{amp},
\end{equation}
where $\delta$ is a balance factor.

\subsubsection{Lightness Perturbation}\label{sec:lightness}
As previously discussed, in order to automatically identify lightness-relevant channels during training, we need to perturb the lightness component of input images. These perturbations should capture the essence of lightness adaptation, while avoiding interfering with other image components such as structural information. To achieve this, we propose a frequency-based perturbation scheme that linearly interpolates the amplitudes of two images, since amplitude information contains more lightness information~\cite{huangdeep2022} that can prevent augmentation artifacts (Fig.~\ref{fig:frequency_aug}).

\begin{figure*}[tbp]
\footnotesize
\begin{center}
\includegraphics[width=1\textwidth]{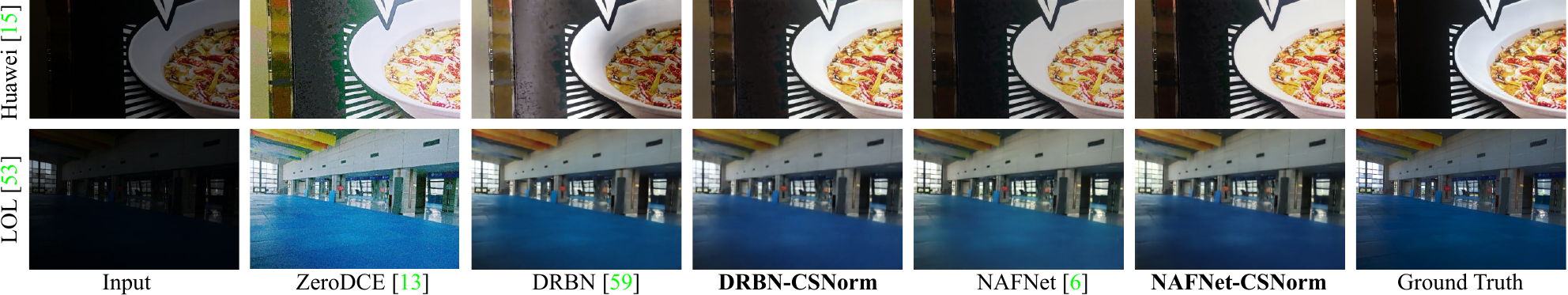}
\end{center}
\vspace{-0.2in}
\caption{Visual results of generalized low-light image enhancement on Huawei~\cite{hai2021r2rnet} and LOL~\cite{wei2018deep} datasets. The models are trained on one dataset (LOL~\cite{wei2018deep} or Huawei~\cite{hai2021r2rnet} dataset) and tested on the other (Huawei~\cite{hai2021r2rnet} or LOL~\cite{wei2018deep} dataset). Equipped with our CSNorm, the generalization abilities of base networks (DRBN~\cite{yang2020fidelity} and NAFNet~\cite{chen2022simple}) are significantly improved. }
\label{fig:lowlight}
\vspace{-0.1cm}
\end{figure*}

\begin{table*}[!t]
\caption{Quantitative results of low-light image enhancement methods on synthetic lightnesses in terms of PSNR and SSIM.}
\label{tab:lowlight}
\vspace{-0.2cm}
\centering
    \renewcommand\arraystretch{1.1}
\resizebox{\textwidth}{!}{%
\begin{tabular}{c|cccc|cccc}
\hline
\multirow{2}{*}{Method} & \multicolumn{4}{c|}{LOL~\cite{wei2018deep}} & \multicolumn{4}{c}{Huawei~\cite{hai2021r2rnet}} \\\cline{2-9}
 & original &interp & scale & average & original &interp & scale & average \\\cline{1-9}
CLANE~\cite{reza2004realization} & 12.77/0.5703 &	13.56/0.5852 &	13.39/0.5806 &	13.24/0.5787 & 13.29/0.4399 & 17.43/0.7220 &	14.51/0.4202 &	15.08/0.5274 \\
LIME~\cite{guo2016lime} & 17.18/0.6130 &	17.49/0.9028 &	17.47/0.8188 &	17.38/0.7782 &	17.09/0.5932 &	15.03/0.7860 &	16.78/0.5778 &	16.30/0.6523 \\
RetinexNet~\cite{wei2018deep} & 16.77/0.5393 &	15.02/0.7870 &	16.33/0.5043 &	16.04/0.6102 &	16.65/0.5010 &	13.12/0.5282 &	16.54/0.4693 &	15.44/0.5495 \\
ZeroDCE~\cite{guo2020zero} & 15.47/0.6521 &	16.35/0.7432 &	15.96/0.6747 &	15.92/0.6900 &	12.53/0.4215 &	8.26/0.5539 &	11.33/0.4022 &	10.70/0.4592 \\

LLFlow~\cite{wang2022low} & 24.93/0.8922 & 18.32/0.9177 &  23.12/0.8901 & 22.12/0.9000 & 20.11/0.6645 & 12.97/0.7122  & 18.11/0.6381 & 17.06/0.6716 \\\hline

SID~\cite{chen2018learning} & 20.52/0.8382 &	15.12/0.7994 &	19.47/0.8305 &	18.37/0.8227 &	18.64/0.6251 &	12.15/0.6687 &	18.05/0.6011 &	16.28/0.6316 \\
SID-CSNorm& \textbf{21.11/0.8312} &	\textbf{16.63/0.8130} &	\textbf{20.80/0.8341} &	\textbf{19.52/0.8194} &	\textbf{18.86/0.6128} &	\textbf{12.77/0.6508} &	\textbf{18.23/0.6114} &	\textbf{16.62/0.6250} \\\hline

DRBN~\cite{yang2020fidelity} & 20.75/0.8426 &	17.43/0.8745 &	20.15/0.8673 &	19.44/0.8614 &	18.80/0.6449 &	11.73/0.6797 &	17.58/0.6303 &	16.04/0.6516 \\
DRBN-CSNorm & \textbf{21.05/0.8533} &	\textbf{18.52/0.8822} &	\textbf{20.64/0.8737} &	\textbf{20.09/0.8697} &	\textbf{18.81/0.6465} &	\textbf{14.87/0.7285} &	\textbf{18.74/0.6383} &	\textbf{17.47/0.6711} \\\hline

{NAFNet}~\cite{chen2022simple} & 23.02/0.8498 &	17.21/0.8733 &	20.22/0.8702 &	19.48/0.8846 &	19.07/0.6483 &	12.53/0.6822 &	17.75/0.6125 &	16.45/0.6476 \\
{NAFNet-CSNorm} & \textbf{23.10/0.8544} &	\textbf{18.75/0.8836} & \textbf{20.97/0.8767} &	\textbf{20.27/0.6476} &	\textbf{19.26/0.6492} &	\textbf{15.02/0.7325} &	\textbf{18.81/0.6410} &	\textbf{17.69/0.6742} \\
\hline
\end{tabular}%
}
\end{table*}

Taking the low light enhancement task as an example, we define the low light and normal light images as $o_l$ and $o_{norm}$, and their Fourier representations as $O_l$ and $O_{norm}$. We linearly combine the amplitude components of $O_l$ and $O_{norm}$ as 
\begin{equation}
    \mathcal{A}(\hat{O}_l)=\lambda \mathcal{A}(O_l) +(1-\lambda)\mathcal{A}(O_{norm}),
\end{equation}
where $\mathcal{A}$ represents amplitude information and $\lambda \in [0,1]$ is randomly sampled. Then the perturbed image $\hat{o}$ is reconstructed through an inverse Fourier transformation $\mathcal{F}^{-1}$ as $\hat{o}=\mathcal{F}^{-1}(\mathcal{A}(\hat{O}_l), \mathcal{P}(O_{l}))$, where $\mathcal{P}$ is phase information.

As shown in Fig.~\ref{fig:frequency_aug}, our frequency-based perturbation mitigates the influence of other factors in the image, such as structure and noise, and focuses more on the lightness itself. The perturbed and original images are used as inputs for different training steps to optimize CSNorm, enabling CSNorm to purposely select the lightness-relevant channels thereby enhancing the network's generalization ability.

\vspace{-0.1cm}
\section{Experiments}
\label{sec:Experiments}
\vspace{-0.1cm}

We do comprehensive evaluations on low-light image enhancement, inverse tone mapping, and image retouching to demonstrate the efficacy of our CSNorm. 

\subsection{Low-light Image Enhancement}\label{sec:lowlight}

\begin{table}[!t]
\caption{Quantitative results of low-light image enhancement methods across datasets in terms of PSNR and SSIM.}
\label{tab:lowlight_real}
\vspace{-0.2cm}
\centering
\footnotesize

    \renewcommand\arraystretch{1.05}
    \setlength{\tabcolsep}{5pt}
\begin{tabular}{ccccccccc}
\hline
\multirow{1}{*}{Train / Test} & \multicolumn{1}{c}{Huawei / LOL} & \multicolumn{1}{c}{LOL / Huawei} \\\hline
CLANE~\cite{reza2004realization} & 10.25/0.5602 & 11.11/0.4152 \\
LIME~\cite{guo2016lime} & 15.25/0.7994 & 15.20/0.5321 \\
RetinexNet~\cite{wei2018deep}	& 15.35/0.5102 & 15.32/0.4855 \\
ZeroDCE~\cite{guo2020zero}  & 15.01/0.5974 & 12.25/0.4194 \\\hline
SID~\cite{chen2018learning} & 17.93/0.7159 & 16.10/0.5689 \\
SID-CSNorm  & \textbf{18.33/0.7725} & \textbf{17.31/0.6105} \\\hline
DRBN~\cite{yang2020fidelity} & 18.10/0.8033 & 15.21/0.5477 \\
DRBN-CSNorm  & \textbf{18.65/0.8105} & \textbf{17.42/0.6122}  \\\hline
NAFNet~\cite{chen2022simple} & 19.05/0.7901 & 17.02/0.6002 \\
NAFNet-CSNorm  & \textbf{19.63/0.8322} & \textbf{17.53/0.6257} \\\hline
\end{tabular}%
\vspace{-0.05in}
\end{table}

\paragraph{Settings.}\label{sec:exp_low_setting}
We conducted experiments on the Huawei~\cite{hai2021r2rnet} and LOL~\cite{wei2018deep} datasets. 
Representative methods such as CLANE~\cite{reza2004realization}, LIME~\cite{guo2016lime}, RetinexNet~\cite{wei2018deep}, LLFlow\cite{wang2022low}, and ZeroDCE~\cite{guo2020zero} are used for comparison. We select SID~\cite{chen2018learning}, DRBN~\cite{yang2020fidelity}, and NAFNet~\cite{chen2022simple} as base networks and integrate our CSNorm into them. The peak signal-to-noise ratio (PSNR) and structural similarity index measure (SSIM) are used as evaluation metrics. 

We conduct experiments in synthetic and realistic settings.
For the synthetic setting, we simulate two input lightness conditions, i.e., interp: the input image is interpolated by the original low-light and ground-truth images (weight is $0.5$), and scale: the input image $x$ is manipulated as $x'=\Lambda x^{\eta}$ ($\eta=1.1, \Lambda=1.2$). For the real-world setting, we use LOL~\cite{wei2018deep} and Huawei~\cite{hai2021r2rnet} for cross-validation. The model is trained on one dataset and tested on the other.

\begin{figure*}[tbp]
\footnotesize
\begin{center}
\includegraphics[width=1\textwidth]{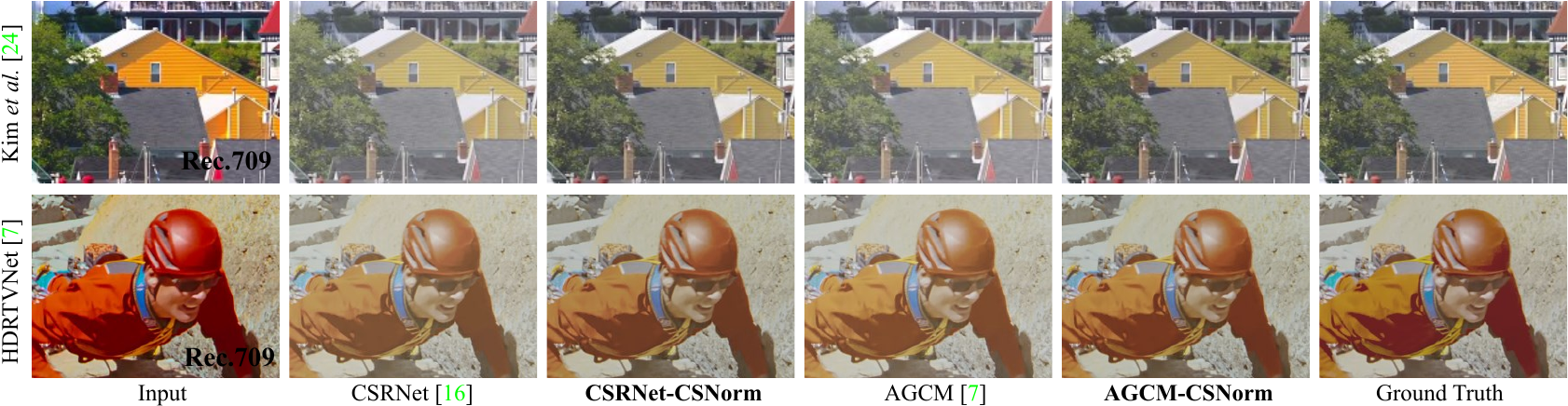}
\end{center}
\vspace{-0.2in}
\caption{Visual results of the generalized inverse tone mapping on Kim \emph{et al.}~\cite{kim2019deep} and HDRTVNet~\cite{chen2021new} datasets. The models are trained on one dataset (Kim \emph{et al.}~\cite{kim2019deep} or HDRTVNet~\cite{chen2021new} dataset) and tested on the other (HDRTVNet~\cite{chen2021new} or Kim \emph{et al.}~\cite{kim2019deep} dataset). The colors of results seem light-colored since they are visualized in the standard Rec.2020 color space.}\label{fig:itm_real}
\vspace{-0.1cm}
\end{figure*}

\begin{table*}[!t]
\caption{Quantitative results of inverse tone mapping methods on synthetic lightnesses in terms of PSNR and SSIM.}
\label{tab:itm_syn}
\vspace{-0.25cm}
\centering
\renewcommand\arraystretch{1}
\resizebox{\textwidth}{!}{
\begin{tabular}{c|cccc|cccc}
\hline
\multirow{2}{*}{Method} & \multicolumn{4}{c|}{Kim \emph{et al.}~\cite{kim2019deep}} & \multicolumn{4}{c}{HDRTVNet~\cite{chen2021new}} \\\cline{2-9}
& original & interp & scale & average  & original & interp & scale & average \\
\hline
CSRNet~\cite{he2020conditional} & {32.22}/0.9472 & 26.74/0.9545 & 27.43/0.9282 & 28.79/0.9433 
& 36.01/0.9717 & 29.22/0.9764 & 27.93/0.9417 & 31.05/0.9632\\
CSRNet-CSNorm  & \textbf{32.53}/\textbf{0.9496} & \textbf{27.14/0.9562} & \textbf{27.83/0.9301} & 
 \textbf{29.16/0.9453} & \textbf{36.15/0.9730} 
& \textbf{29.46/0.9766} & \textbf{28.15/0.9428} & \textbf{31.25/0.9641}\\\hline
AGCM~\cite{chen2021new} & {32.44}/0.9482 & 26.87/0.9551 & 27.51/0.9297 & 28.94/0.9443 & {36.25/0.9733} & 29.36/0.9767 & 27.98/0.9418 &  31.19/0.9639\\
AGCM-CSNorm & \textbf{32.61}/\textbf{0.9501} & \textbf{27.26/0.9573} & \textbf{27.91/0.9322} & \textbf{29.26/0.9465} & \textbf{36.31/0.9734} & \textbf{29.57/0.9781} & \textbf{28.26/0.9433} & \textbf{31.38/0.9649}
 \\\hline
\end{tabular}
}
\end{table*}

\vspace{-0.5cm}
\paragraph{Results.} 
Table~\ref{tab:lowlight} compares the performance of our proposed method (name with \textbf{-CSNorm}) and baseline methods on Huawei and LOL datasets. While previous methods achieve good results on original low-light conditions, they have a poor generalization ability on unknown lightness conditions. In contrast, our methods exhibit superior generalization ability, outperforming corresponding backbones by over 0.6dB on the two datasets. Our CSNorm also keeps the performance on original low-light images. Note that our objective is to improve the initial network rather than achieve state-of-the-art performance.

Table~\ref{tab:lowlight_real} shows generalization performance across different datasets. 
It can be seen that, previous methods tend to overfit the training dataset and have poor abilities for generalization. Our CSNorm improves the performance of all base networks, which greatly enhances their generalization abilities on unknown lightness conditions. We also show qualitative results in Fig.~\ref{fig:lowlight}. Even though there is a large discrepancy between training and testing images (two datasets are separately captured with different lightnesses), the base networks equipped with CSNorm produce visually pleasing results on unknown lightness conditions.

\subsection{Inverse Tone Mapping}
\paragraph{Settings.}
We perform inverse tone mapping using CSRNet~\cite{he2020conditional} and AGCM~\cite{chen2021new} as base networks, where we deepen the AGCM We use HDRTVNet~\cite{chen2021new} and Kim~\emph{et al.}~\cite{kim2019deep} for training and testing. We conduct experiments on the original, the interpolated, and the scaled lightness conditions same as Sec.~\ref{sec:lowlight}. 

\begin{table}[t]
\caption{Quantitative results of inverse tone mapping methods using realistic datasets in terms of PSNR and SSIM.}
\label{tab:itm_real}
\vspace{-0.25cm}
\centering
\renewcommand\arraystretch{1.05}
\setlength{\tabcolsep}{5pt}
\resizebox{0.43\textwidth}{!}{%
\begin{tabular}{ccccccccc}
\hline
\multirow{1}{*}{Train / Test} & \multicolumn{1}{c}{Kim \emph{et al.} / HDRTVNet} & \multicolumn{1}{c}{HDRTVNet / Kim \emph{et al.}} \\\hline
CSRNet~\cite{he2020conditional} & 33.47/0.9642 & 31.06/0.9501 \\
CSRNet-CSNorm  & \textbf{34.02/0.9677} & \textbf{31.35/0.9523}  \\\hline
AGCM~\cite{chen2021new} & 33.52/0.9651 & 31.22/0.9512 \\
AGCM-CSNorm  & \textbf{34.11/0.9687} & \textbf{31.67/0.9539} \\\hline
\end{tabular}%
} 
\end{table}

\vspace{-0.4cm}
\paragraph{Results.} 
Table~\ref{tab:itm_syn} shows the quantitative results on the synthetic lightness conditions. CSRNet and AGCM have good performances on original SDR frames but perform poorly when the lightness condition changes. In contrast, CSRNet-CSNorm and AGCM-CSNorm, which only add our CSNorm to base networks, achieve well performances on unknown lightness conditions. We also show real-world results in Table~\ref{tab:itm_real} and Fig.~\ref{fig:itm_real}. The results demonstrate that our CSNorm has a strong generalization ability across different lightness conditions. It is worth noting that, CSNorm does not affect the performance of the original SDR frames, which confirms that the selected channels only affect the lightness-relevant information without altering the overall data distribution.

\begin{figure}[tbp]
\footnotesize
\begin{center}
\includegraphics[width=0.47\textwidth]{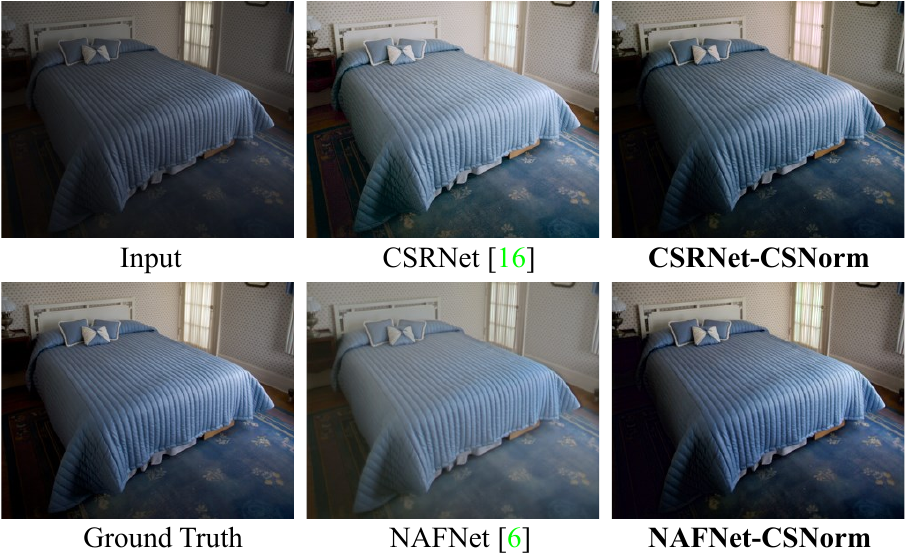}
\end{center}
\vspace{-0.25in}
\caption{Visual comparisons of the generalized image retouching on the MIT-Adobe FiveK~\cite{bychkovsky2011learning} dataset. The models are trained on the original dataset and tested on the scaled lightness condition.}
\label{fig:retouching}
\end{figure}

\subsection{Image Retouching}

\paragraph{Settings.} We adopt MIT-Adobe FiveK~\cite{bychkovsky2011learning} for training and testing.  The experiments are conducted on lightness conditions as Sec.~\ref{sec:lowlight}. We select CSRNet~\cite{he2020conditional},  DRBN~\cite{yang2020fidelity}, and NAFNet~\cite{chen2022simple} as base networks to plug our CSNorm.

\vspace{-0.3cm}
\paragraph{Results.}
We show the quantitative results in Table~\ref{tab:retouching}. It can be seen that, the base network (CSRNet) has a good performance on the original image but suffers from poor generalization ability. Its performance drops about 5dB when lightness changes from the original one to the interpolation one. Based on CSRNet network, our method (CSRNet-CSNorm) remarkably improves the original model's average capability by over 1 dB, which demonstrates CSNorm's powerful generalization ability. We also show the visual results in Fig.~\ref{fig:retouching}.

\begin{table}[t]
\caption{Quantitative results of image retouching methods on synthetic lightnesses in terms of PSNR and SSIM.}
\label{tab:retouching}
\vspace{-0.2cm}
\centering
    \renewcommand\arraystretch{1.1}
    \setlength{\tabcolsep}{4pt}
\resizebox{0.48\textwidth}{!}{%
\begin{tabular}{c|cccc}
\hline
Method  & original & interp & scale & average  \\\hline
DRBN~\cite{yang2020fidelity} & {22.11/0.8622} &	23.54/0.8757 &	20.51/0.8544 &	22.05/0.8641
 \\
DRBN-CSNorm & \textbf{22.43/0.8679} &	\textbf{23.97/0.8792} &	\textbf{20.68/0.8561} &	\textbf{22.36/0.8677 }  \\\hline
CSRNet~\cite{he2020conditional} & \textbf{23.52}/{0.8865} & 24.69/0.8943 & 22.76/0.8651 &	23.65/0.8823
 \\
CSRNet-CSNorm& {23.41}/\textbf{0.8838} &	\textbf{25.34/0.9026} &	\textbf{22.85/0.8676} &	\textbf{23.86/0.8846}  \\\hline
NAFNet~\cite{chen2022simple} & \textbf{23.71/0.8901} & 24.77/0.8966	 & 22.85/0.8685 & 
23.82/0.8850 \\
NAFNet-CSNorm& {23.54/0.8922} &	\textbf{25.40/0.9033} &	\textbf{22.97/0.8690} &	\textbf{24.03/0.8881}  \\\hline

\end{tabular}
}
\end{table}

\begin{table}[t]
    \centering
\normalsize
        \caption{Quantitative results of normalization methods, where the models are trained on the LOL~\cite{wei2018deep} dataset and tested on the Huawei~\cite{hai2021r2rnet} dataset.}
\label{tab:norm}
    \vspace{-0.2cm}
    \renewcommand\arraystretch{1.05}
\setlength{\tabcolsep}{4pt}
        \resizebox{0.48\textwidth}{!}{

    \begin{tabular}{cccccccc}
\hline
Method & DRBN-BN & DRBN-IN & DRBN-BIN & DRBN-CSNorm \\ \hline
PSNR/SSIM & 16.05/0.5537 & 16.98/0.5821 & 17.11/0.5817 & \textbf{17.42/0.6122} \\\hline
\end{tabular}}
\end{table}

\begin{figure}[tbp]
\footnotesize
\begin{center}
  \begin{subfigure}{1\linewidth}
    \includegraphics[width=1\textwidth]{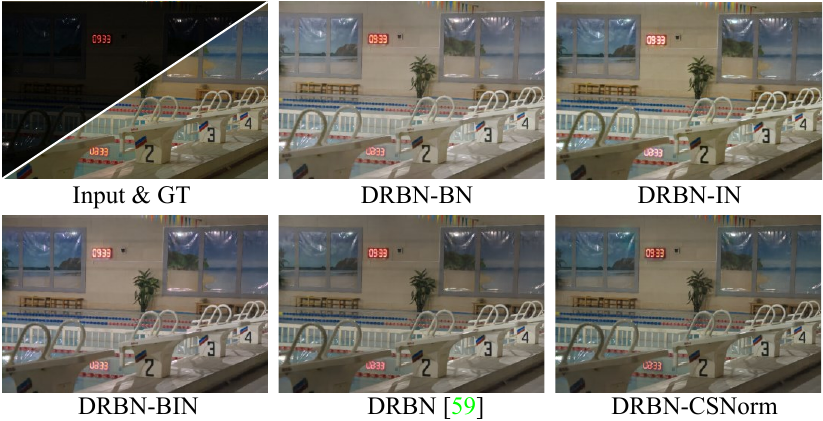}
\caption{The models are trained and tested on the LOL~\cite{wei2018deep} dataset.}    \label{fig:BIN1}
  \end{subfigure}
  \begin{subfigure}{1\linewidth}
    \includegraphics[width=1\textwidth]{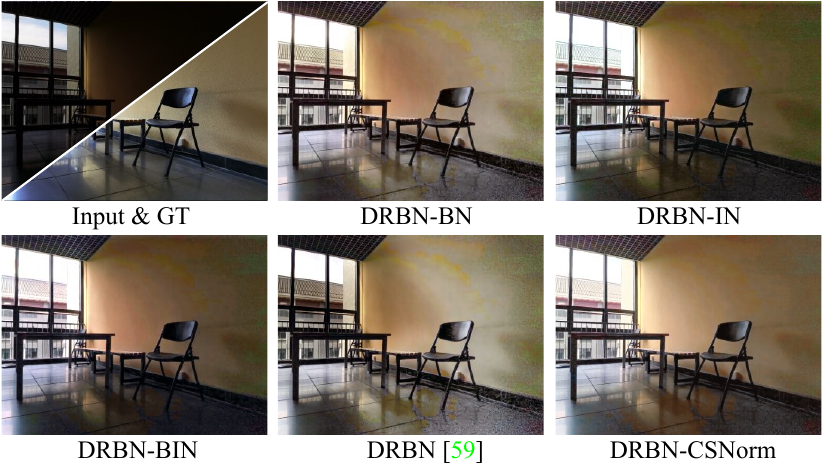}
    \vspace{-0.1in}
\caption{The models are trained on the LOL~\cite{wei2018deep} dataset and tested on the Huawei~\cite{hai2021r2rnet} dataset. }
  \end{subfigure}
\end{center}
\vspace{-0.2in}
\caption{Ablation study on normalization techniques. (a) CSNorm keeps the base network's ability. (b) CSNorm can generalize to other unknown lightness conditions.}
\label{fig:norm}
\vspace{-0.05in}
\end{figure}

\section{Analysis}

\paragraph{Feature visualization.}

\begin{figure}
  \centering
\includegraphics[width=0.47\textwidth]{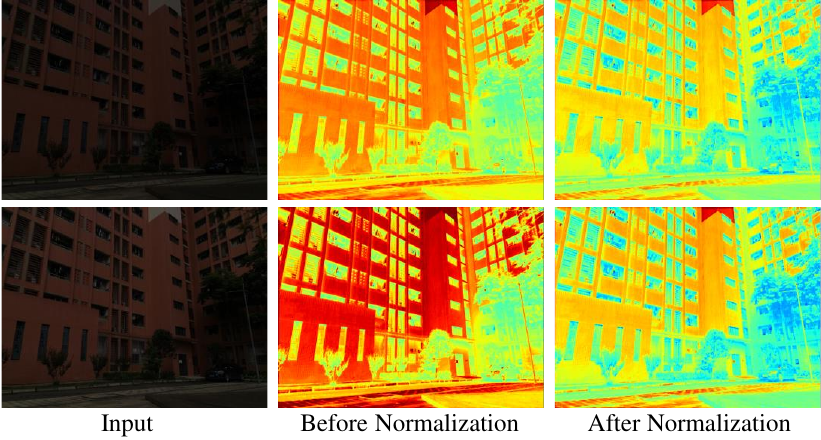}
  \caption{Visualization of lightness-relevant channels. The channels are adaptively selected by our CSNorm.}\label{fig:feat}
\end{figure}

We visualize the selected channels to demonstrate our CSNorm can effectively enhance the model's ability to generalize to different lightnesses. We show the selected channels (\emph{i.e.,} lightness-relevant channels) in Fig.~\ref{fig:feat}. It can be seen that, the channels extracted from different lightness conditions have different characteristics, which may lead to poor generalization ability. Our CSNorm selects this channel and normalizes it, which substantially produces lightness-independent information for generalization.

\vspace{-0.3cm}
\paragraph{Comparisons to other normalizations.}

We compare our CSNorm with conventional normalization techniques, including Batch Normalization (BN)~\cite{ioffe2015batch}, Instance Normalization (IN)~\cite{ulyanov2016instance}, and Batch Instance Normalization (BIN)~\cite{nam2018batch}, by plugging them into DRBN~\cite{yang2020fidelity}. Since the alternating strategy is specially designed for our CSNorm and may be harmful to conventional normalization techniques, we train BN, IN, and BIN only with the data perturbation. We show the quantitative and qualitative results in Table~\ref{tab:norm} and Fig.~\ref{fig:norm}, respectively. Compared with conventional normalization techniques, our CSNorm effectively keeps the performance on the known lightness condition and has a well generalization ability to unknown lightness conditions, which avoids unsatisfied artifacts.

\vspace{-0.3cm}
\paragraph{Training strategy and data perturbation.}
We take ablation studies on training strategy and data perturbation. For the training strategy, we replace the alternating training strategy with the mixed training strategy, where the network is trained by mixing original and perturbed data. For the data perturbation, we replace the frequency-based perturbation with linearly blending the input images and the ground truth images. These aforementioned ablation experiments are conducted on the LOL~\cite{wei2018deep} dataset and tested on the Huawei~\cite{hai2021r2rnet} dataset. As shown in Table~\ref{tab:aba_strategy}, the alternating training strategy gets a better performance compared with mixed training, while our frequency-based data perturbation improves the model's generalization, demonstrating the effectiveness of our design.
Simply training CSNorm with a diverse range of lighting conditions images cannot effectively identify the lightness-relevant channels due to the influence of content, which is detrimental to the lightness generalization. We conduct experiments on the ME dataset~\cite{afifi2021learning} (Table~\ref{tab:R1-0}). It can be seen that, without the alternating training strategy (ATS), the network cannot generalize well to unknown lightness conditions (Exp0 and Exp6 are unknown lower and higher lightness).

\begin{table}[t]
    \caption{Ablation study on training strategy and data perturbation.}
\label{tab:aba_strategy}
\vspace{-0.2cm}
    \centering
    \renewcommand\arraystretch{1.05}
\setlength{\tabcolsep}{4pt}
        \resizebox{0.48\textwidth}{!}{
    \begin{tabular}{cccccccc}
\hline
\multirow{1}{*}{Training strategy} & \multicolumn{2}{c}{Mixed} & \multicolumn{2}{c}{Alternating}\\\cline{1-5}
Data perturbation &       
   Linear & Amp & Linear & Amp (ours) \\\hline
DRBN~\cite{yang2020fidelity} & 14.35/0.5526 & 14.10/0.5382 & - & - \\

DRBN-CSNorm & 16.79/0.5950 & 16.81/0.5972 & 17.15/0.6097 & \textbf{17.42/0.6122} &\\\hline
\end{tabular}}
\end{table}

\vspace{-0.25cm}
\paragraph{Parameters.}\label{sec:param}
Our proposed CSNorm is lightweight that can be plugged into existing networks with nearly no parameter increase, which avoids the heavy storage cost. For example, given a feature with 64 channels, our CSNorm just requires 16.5k parameters to identify and normalize lightness-relevant channels, owing to the gating module and the affine transformation in normalization. The CSNorm's number of parameters grows linearly with the number of channels. 

\begin{table}[t]
    \centering
      \scriptsize
    \caption{Experiment results of exposure correction in terms of PSNR (dB) and SSIM.
  }
  \label{table:1}
  \vspace{-0.2cm}
  \renewcommand\arraystretch{1.0}
    \begin{tabular}{c|c|c|ccc}
\hline
       Method & ME~\cite{afifi2021learning} & Exp0 & Exp6 \\ \hline
       DRBN~\cite{yang2020fidelity} & 19.65/0.8292 & 18.10/0.7843 & 13.43/0.7339\\
       DRBN w/o ATS & 21.31/0.8345 & 18.59/0.7321 & 15.47/0.7917\\
       DRBN w/ ATS & \textbf{21.63/0.8396} & \textbf{18.98/0.7673} & \textbf{16.31/0.7928}\\ 
\hline
    \end{tabular}
    \label{tab:R1-0}
\end{table}

\begin{table}[t]
  \caption{Experiments on lightness-only datasets.
  }
  \vspace{-0.7cm}
  \label{table:R3}
    \begin{center}
    \resizebox{1\linewidth}{!}{
\begin{tabular}{cccccc}
\hline
Train / Test  & LOL / LOL &  LOL / Huawei & Huawei / Huawei & Huawei / LOL  \\ \hline
DRBN~\cite{yang2020fidelity} & 20.98/0.8922 & 15.13/0.5778 & 18.94/0.6517  & 18.76/0.8190  \\
DRBN-CSNorm & \textbf{24.33/0.8998} & \textbf{17.40/0.5883} & \textbf{19.10/0.6505}  & \textbf{19.15/0.8255}  \\ 
\hline
\end{tabular}
}
\end{center}
  \vspace{-0.6cm}

\end{table}

\vspace{-0.25cm}
\paragraph{Evaluation on known lightness-only datasets.}

To further demonstrate the effectiveness of our CSNorm, we conduct experiments on lightness-only datasets. We transform the color image into Ycbcr color space and use the Y channel since it represents the luminance or lightness information. 
Experiment results in Table~\ref{table:R3} show that our method can effectively enhance generalization ability without sacrificing the discrimination of the features. 

\vspace{-0.25cm}
\paragraph{Amplitude-related information.}
Amplitude-related information has been proven related to the lightness components in previous works~\cite{huangdeep2022}. We implicitly utilize the amplitude-related information as a detailed lightness perturbation manner in the alternating training strategy, enabling CSNorm to identify the lightness-related channels. Thus, this amplitude-based lightness perturbation is orthogonal to CSNorm and its format is not introduced into CSNorm.
Note that other lightness perturbation manner can also drive CSNorm's training, \textit{e.g.}, linear interpolation in Table~\ref{tab:aba_strategy}, while our adopted amplitude-based lightness perturbation experimentally achieves higher performance.

\section{Conclusion and Discussion}

In this work, we propose CSNorm, a novel normalization technique customized for generalized lightness adaptation. It purposely normalizes lightness-relevant channels while keeping other channels unchanged, which empowers existing lightness adaptation methods with the generalization ability to wide-range lightness scenes. Except for the sufficient generalization ability on the unknown lightnesses, CSNorm keeps the reconstruction accuracy on the known lightness. The proposed CSNorm is architecture-agnostic which we validate, making it simple, lightweight, and plug-and-play. Extensive experiments on multiple tasks and benchmark datasets verify the effectiveness of our proposed CSNorm to enhance the generalization of existing lightness adaptation methods. We believe our method would inspire more valuable generalization methods for lightness adaptation, and  holds potential for application in other tasks.  

Despite the promising preliminary results, there are still some real-world conditions that have not been considered in this paper. For images with no discernible details or high-level noise levels, it is indeed challenging for our CSNorm due to the loss of necessary information. However, our lightweight and plug-and-play design may enable it to handily collaborate with other methods, either by inserting or by following inpainting or denoising networks. Moreover, we believe there is great potential to explore more robust data perturbation methods, e.g., Retinex model-based lightness decomposition, to further uncover the capabilities of our method.

\section*{Acknowledgments}
This work was supported in part by the National Natural Science Foundation of China under Grants 62131003 and 62021001.

\medskip

{\small
\bibliographystyle{ieee_fullname}
\bibliography{egbib}
}

\end{document}